\documentclass{article}

\usepackage[preprint, nonatbib]{neurips_2022}

\usepackage[utf8]{inputenc} 
\usepackage[T1]{fontenc}    
\usepackage{hyperref}       
\usepackage{url}            
\usepackage{booktabs}       
\usepackage{amsfonts}       
\usepackage{nicefrac}       
\usepackage{microtype}      
\usepackage{xcolor}         

\usepackage{graphicx}
\usepackage{algorithm}
\usepackage{algorithmic}

\usepackage{amssymb}
\usepackage{amsmath}

\usepackage{wrapfig}

\usepackage[belowskip=-15pt,aboveskip=5pt]{caption}
\graphicspath{{figs/}}  

\usepackage[backend=biber,natbib=true, sorting=none]{biblatex}

\title{Multi-Modal Fusion by Meta-Initialization}

\author{
    Matthew T.~Jackson\thanks{Equal Contribution} \\
    Department of Engineering Science \\
    University of Oxford \\
    \texttt{jackson@robots.ox.ac.uk} \\
    \And
    Shreshth A.~Malik\textsuperscript{*} \\
    Department of Engineering Science \\
    University of Oxford \\
    \texttt{shreshth@robots.ox.ac.uk} \\
    \And
    Michael T.~Matthews \\
    Department of Computer Science \\
    University College London \\
    \And
    Yousuf Mohamed-Ahmed \\
    Department of Computer Science \\
    University College London \\
}

\begin{document}

\maketitle

\begin{abstract}

    When experience is scarce, models may have insufficient information to adapt to a new task.
    In this case, auxiliary information---such as a textual description of the task---can enable improved task inference and adaptation.
    In this work, we propose an extension to the Model-Agnostic Meta-Learning algorithm (MAML), which allows the model to adapt using auxiliary information as well as task experience.
    Our method, Fusion by Meta-Initialization (FuMI), conditions the model initialization on auxiliary information using a hypernetwork, rather than learning a single, task-agnostic initialization.
    Furthermore, motivated by the shortcomings of existing multi-modal few-shot learning benchmarks, 
    we constructed iNat-Anim---a large-scale image classification dataset with succinct and visually pertinent textual class descriptions. On iNat-Anim, FuMI significantly outperforms uni-modal baselines such as MAML in the few-shot regime. 
    The code for this project and a dataset exploration tool for iNat-Anim are publicly available at 
    \url{https://github.com/s-a-malik/multi-few}.

\end{abstract}


\section{Introduction}
\label{sec:intro}



Learning effectively in resource-constrained environments is an open challenge in machine learning \cite{maml, protonets, wang2020generalizing}. Yet humans are capable of rapidly learning new tasks from limited experience, in part by drawing on auxiliary information about the task. 
This information can be particularly helpful in the few-shot regime, as it can highlight features that have not been seen directly in task experience, but are necessary to solve the task. For example, Figure \ref{figure:inat-anim} shows an example image classification task where a text description of the class contains discriminative information that is not contained in the training (support) images.
Designing algorithms that can incorporate auxiliary information into meta-learning approaches has consequently attracted much attention \cite{ma2022multimodality, ma2021smil, am3, akata2015label, WangYWDG19, TsaiHS17, SchonfeldESDA19}.

Model-agnostic meta-learning (MAML) \cite{maml} is a popular method for few-shot learning. However, it cannot incorporate auxiliary task information. 
In this work, we propose Fusion by Meta-Initialization (FuMI), an extension of MAML which uses a hypernetwork \cite{hypernetworks} to learn a mapping from auxiliary task information to a parameter initialization. While MAML learns an initialization that facilitates rapid learning across all tasks, FuMI conditions the initialization on the specific task to enable improved adaption.


Existing multi-modal few-shot learning benchmarks largely rely on hand-crafted feature vectors for each class \cite{awa, WahCUB_200_2011}, or use noisy language descriptions from sources such as Wikipedia \cite{Paz-ArgamanTCA20,Elhoseiny17}. For this reason, we release iNat-Anim---a large animal species image classification dataset with high quality descriptions of visual features.
On this benchmark, we find that FuMI significantly outperforms MAML in the very-few-shot regime.

\section{Background}
\label{sec:bg}

\begin{figure}[t]
    \centering
    \includegraphics[width=\linewidth]{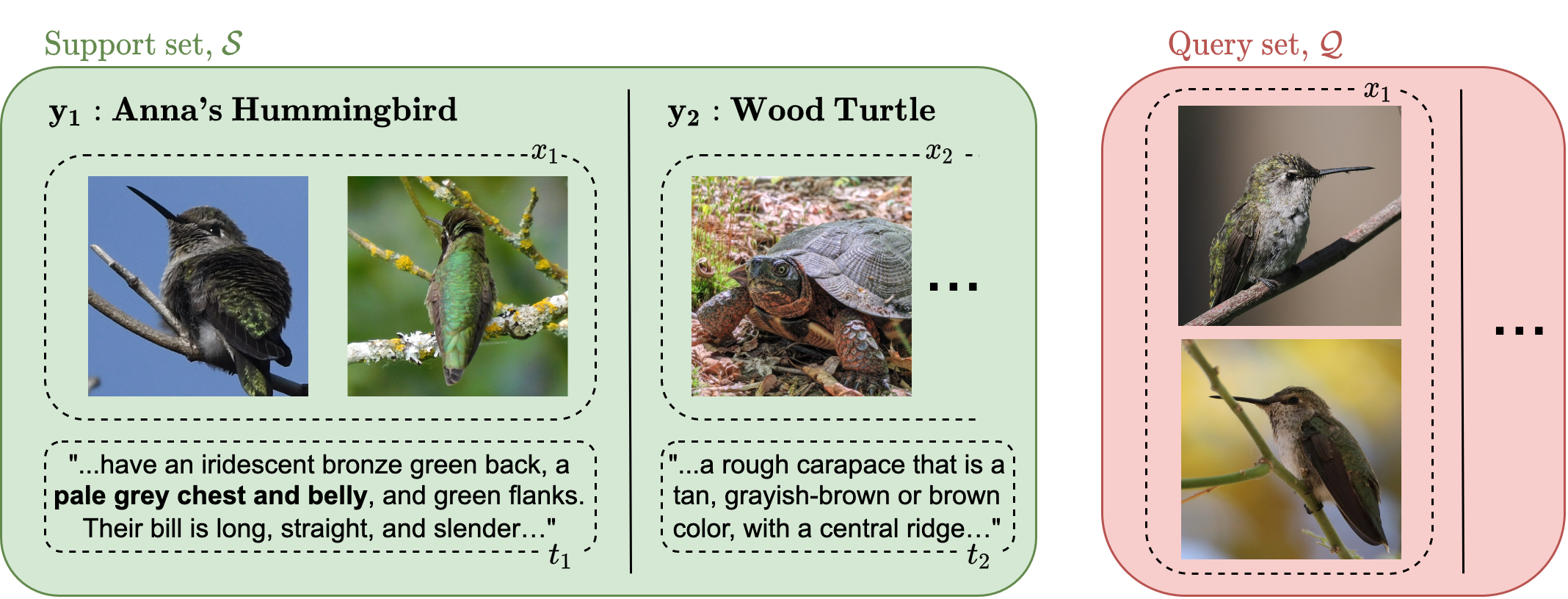}
    \caption{An example few-shot learning task, using images and class descriptions from our proposed dataset, iNat-Anim. Here, we see the class description contains information (the colour of the bird's breast) which is not found in the class images (as they are all turned away). 
    }
    \label{figure:inat-anim}
\end{figure}

In the meta-learning framework \cite{maml}, we suppose tasks are drawn from a task distribution $p(\mathcal{T})$. At meta-train time, the model $f_{\theta}$ is evaluated on a series of tasks $\mathcal{T}_i \in \mathcal{D_{\text{train}}}$, where $\mathcal{D_{\text{train}}}$ is a finite set of samples from $p(\mathcal{T})$. This gives task loss $\mathcal{L}_{\mathcal{T}_i}$, which is used to update the model parameters $\theta$ in accordance with the meta-learning algorithm. At meta-test time, the trained model is evaluated on all tasks in $\mathcal{D_{\text{test}}}$, another set of samples from $p(\mathcal{T})$.

In an $N$-shot, $K$-way multi-modal classification problem\footnote{For consistency with our dataset, the problem setting formulation is for classification. However our method can also be applied to regression and reinforcement learning.}, a task $\mathcal{T} = (\mathcal{S}, \mathcal{Q})$ is defined by a support set $\mathcal{S} = \{(\{x_{i,j}\}_{j=1}^{N}, t_i, y_i)\}_{i=1}^{K}$ and a query set $\mathcal{Q} = \{(\{x_{i,j}\}_{j=1}^{M}, y_i)\}_{i=1}^{K}$, where $M$ is the number of query shots.
The support set contains $N$ samples and auxiliary class information $t_i$ for each of the $K$ classes, which are used by the meta-learner
to train an adapted model. Once this has been trained, the adapted model is evaluated on the unseen query set, giving task loss $\mathcal{L}_{\mathcal{Q}}$. In the context of our work, $t_i$ denotes the textual description of the class $y_i$, meaning each class has a textual description and $N$ support images. Figure \ref{figure:inat-anim} shows an example task using the notation outlined here.


\section{Data}
\label{sec:data} 

\paragraph{Existing Multi-Modal Few-shot Benchmarks.} 
While there are a number of popular uni-modal few-shot learning benchmarks \cite{vinyals2016matching, lake2015human, van2018inaturalist}, multi-modal benchmarks are less common. Some works simply extend few-shot benchmarks by using the class label as auxiliary information \cite{am3, schwartz2022baby}.  Benchmarks explicitly incorporating auxiliary modalities include Animals with Attributes (AWA) \cite{awa} and Caltech-UCSD-Birds (CUB) \cite{WahCUB_200_2011} which augment images of animals/birds with hand-crafted class attributes. While semantic class features can be highly discriminative, they require manual labelling and are thus difficult to obtain at scale. 
Recent work instead uses the more general approach of using natural language descriptions, for example, through augmenting CUB with Wikipedia articles \cite{Paz-ArgamanTCA20,Elhoseiny17}. However, these articles are subject to change and visual information is sparse, thus reducing the relative benefit of the auxiliary information. 







\paragraph{The iNat-Anim Dataset.}
Motivated by these shortcomings, we constructed the \textit{iNat-Anim}\footnote{\url{https://doi.org/10.5281/zenodo.6703088}} dataset. iNat-Anim consists of 195,605 images across 673 animal species, which is orders of magnitude larger than existing benchmarks (AWA and CUB). The images are a subset of the iNaturalist 2021 CVPR challenge \cite{inat21} and have been augmented with textual descriptions from Animalia \cite{animalia} to provide auxiliary information about each species. The descriptions are typically short and are qualitatively pertinent to the visual characteristics of the animal (Figure \ref{figure:inat-anim}).
See Appendix \ref{sec:appdata} for further details. 


\section{Method}




\setlength{\intextsep}{0pt}%
\begin{wrapfigure}{R}{0.5\textwidth}
\begin{minipage}{0.5\textwidth}
\begin{algorithm}[H]
\caption{FuMI for few-shot classification, with differences from MAML in \textcolor{red}{red}.}
\label{alg:fumi}
\begin{algorithmic}
\REQUIRE $p(\mathcal{T})$: distribution over tasks
\REQUIRE $\alpha, \beta$: step size hyperparameters \\
Randomly initialize $\theta^{\text{Body}}, \textcolor{red}{\phi}$
\WHILE{not done}
    \STATE{Sample task $(\mathcal{S}, \mathcal{Q}) \sim p(\mathcal{T})$
    \textcolor{red}{\FORALL{class information $t_i$ in $\mathcal{S}$}
        \STATE{$\theta^{\text{Head}}_i = g_{\phi}(t_i)$}
    \ENDFOR \\
    $\theta = (\theta^{\text{Head}}, \theta^{\text{Body}})$} \\
    Adapt parameters $\theta' = \theta - \alpha \nabla_{\theta}\mathcal{L}_{\mathcal{S}}(f_{\theta})$ \\
    Update network body initialization \\
    $\theta^{\text{Body}} \leftarrow \theta^{\text{Body}} - \beta \nabla_{\theta^{\text{Body}}}\mathcal{L}_{\mathcal{Q}}(f_{\theta'})$ \\
    \textcolor{red}{Update hypernetwork \\
    $\phi \leftarrow \phi - \beta \nabla_{\phi}\mathcal{L}_{\mathcal{Q}}(f_{\theta'})$}}
\ENDWHILE
\end{algorithmic}
\end{algorithm}
\end{minipage}
\end{wrapfigure}

We propose Fusion by Meta-Initialization (FuMI): a gradient-based model for multi-modal few-shot learning. This model extends MAML by conditioning the meta-initialization of task-specific model parameters on their associated task information, thereby incorporating the auxiliary information into the tuned model.





Suppose we are training a neural network for $K$-way classification, with a fully-connected final layer (head). The parameters of the final layer $\theta^{\text{Head}}$ can be partitioned such that each $\theta^{\text{Head}}_i$ generates the class probability density $p(c_i | x)$ for a particular class $c_i$. MAML learns an initialization $\theta = (\theta^{\text{Head}}, \theta^{\text{Body}})$ and updates it with gradient descent in the inner-loop, uninformed by the auxiliary task information $t$. However, given $t$, we may instead condition the initialization of each class head $\theta^{\text{Head}}_i$ on the auxiliary information for its associated class $t_i$, thereby generating a class-specific initialization.


In FuMI (Algorithm \ref{alg:fumi}), we use a hypernetwork $g_{\phi}$ \cite{hypernetworks} to generate this initialization for the final layer, by computing $\theta^{\text{Head}}_i = g_{\phi}(t_i)$ for each class description $t_i$. As in MAML, a shared initialization $\theta^{\text{Body}}$ is used for the remainder of the network. All network weights $\theta$ are then tuned by gradient descent in the inner loop, giving $\theta'$. In the outer loop, the query set loss $\mathcal{L}_{\mathcal{Q}}(f_{\theta'})$ is used to update both the network body initialization $\theta^{\text{Body}}$ and hypernetwork parameters $\phi$.

\section{Experiments}
\label{sec:exp}


\subsection{Setup}
\label{sec:setup}

\paragraph{Experimental set-up} The multi-modal few-shot problem is described in Section \ref{sec:bg}. We evaluated 5-way classification accuracy on iNat-Anim with up to 10 shots. We report the average meta-test accuracy for each model across 5 random seeds. The meta-test split consisted of 1,000 randomly sampled tasks where all classes in this split had not previously been seen in training. Each test task had 20 randomly-sampled query images from each class, ensuring there was no dataset imbalance. 

\paragraph{Baselines}

We compared few-shot learning performance of FuMI to MAML as a natural uni-modal baseline. We additionally compare performance to metric-based meta-learning approaches: 1) Prototypical Networks \cite{protonets}, which computes a mean image embedding (prototype) for each class and classifies query images as the class corresponding to the closest prototype by Euclidean distance, 
2) AM3 \cite{am3}, a multi-modal extension to Prototypical Networks, which learns an adaptable convex combination of the image prototype with another prototype computed from the auxiliary modality. 
We used the same pre-trained image and text encoders (BERT \cite{bert} and ResNet-152 \cite{he2016deep}) for all models to enable fair comparison across methods. Appendix \ref{sec:hypers} discusses implementation details.

\subsection{Results}
\label{sec:results}

\begin{wrapfigure}{r}{0.5\linewidth}
    \centering
        \includegraphics[width=\linewidth]{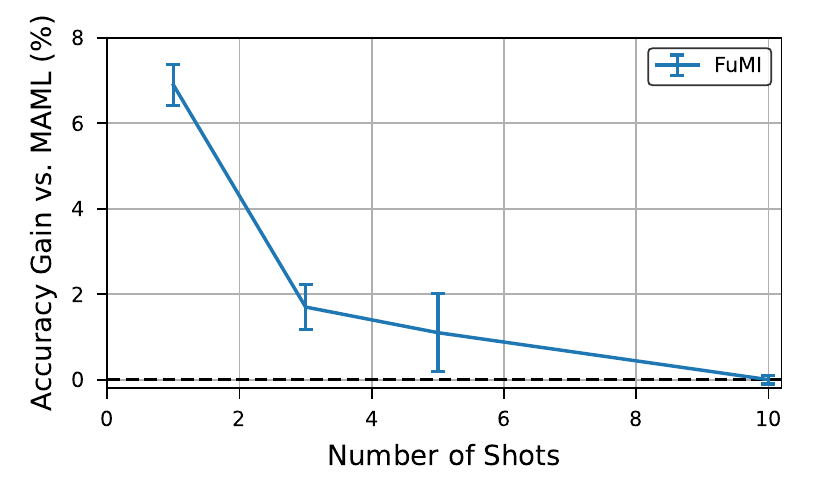}
        \caption{Few-shot classification accuracy of FuMI on iNat-Anim compared to MAML. Error bars show the uncertainty across 5 random seeds. 
        }
        \label{figure:fumi-vs-maml}
        \vspace{-5pt}
\end{wrapfigure}

\paragraph{Multi-modal fusion improves performance in the very-few-shot regime.} Figure \ref{figure:fumi-vs-maml} shows the relative performance gain of FuMI compared to MAML. We find that using the task-specific initialization provides significant improvements given very limited task data, whilst performance is similar with additional examples. This is as expected, since the relative information gain from auxiliary information is greater when there are fewer examples per class.

\paragraph{Metric-based methods outperform gradient-based methods on iNat-Anim.}
Table \ref{tab:results} shows the results for FuMI against the other uni- and multi-modal baselines. We find that FuMI under-performs compared to the other meta-learning approaches. We note that gradient-based meta-learning models can be particularly sensitive to hyperparameters. Metric-based approaches were observed to be more robust on our dataset.


\begin{table}[t]
    \centering
    \caption{Few-shot classification accuracy for uni-modal (top) and multi-modal (bottom) models on iNat-Anim  over 5 random seeds.}
    \begin{tabular}{@{}l l c c c c c@{}} 
     \toprule
     \textbf{Model} & \multicolumn{5}{c}{\textbf{5-way Test Accuracy (\%)}} \\
     \toprule & 0-shot  & 1-shot & 3-shot & 5-shot & 10-shot \\
     \midrule 
     \midrule
     Proto. Nets \cite{protonets} & - & 71.7(2) & 83.9(3) & 85.9(2) & 88.3(2) \\
     MAML \cite{maml} & - & 72(1) &  81(1) & 84(2) & 87.1(1) \\
     \midrule 
     AM3 \cite{am3} & 71.0(8) & 80.8(4) & 85.9(5) & 86.3(6) & 88.5(2) \\
     FuMI (ours)  & - & 78.9(4) & 82.7(6) & 85.1(4) & 87.1(2) \\
     \bottomrule
    \end{tabular}
    \label{tab:results}
\end{table}

\section{Related Work}
\label{sec:related-work}

A range of other MAML extensions have been recently proposed, with improvements including training stability \cite{mamltrain}, avoiding computational overhead from second-order derivatives \cite{esmaml, reptile}, and exploration in meta-reinforcement learning \cite{emaml}.
\citet{mmaml} use the entire uni-modal support set (rather than auxiliary task information) to modulate the initialization of the entire network.
\citet{rlvfr} find no decrease in performance when updating only the network head in the inner loop, thereby concluding that the features learned in the network body are directly reused across tasks.
Based on this,
in addition to early experimentation, 
we use the hypernetwork to directly initialize only the network head in FuMI.
An alternative approach to few-shot learning is metric-based meta-learning \cite{protonets, am3, siamese, relation},
which we evaluate on iNat-Anim in Section \ref{sec:exp}.







\section{Conclusions}
\label{sec:conc}

\paragraph{Contributions} In this work, we have introduced Fusion by Meta-Initialization, a multi-modal gradient-based meta-learning algorithm. FuMI significantly outperforms MAML baselines given very limited data, highlighting the effectiveness of auxiliary information on few-shot performance. 
To fill the need for large-scale benchmarks, we also constructed iNat-Anim, a few-shot image classification dataset with high-quality class descriptions. 
We hope that this will enable further work on the intersection of meta-learning and multi-modal models.

\paragraph{Limitations and Further Work} Methodologically, we note that the gradient-based inner-loop of FuMI makes it vulnerable to catastrophic forgetting of the auxiliary information used for initialization. Insights from continual learning 
could help mitigate against this \cite{de2021continual}. Experimentally,
we plan to broaden our evaluation of FuMI to further image-text few-shot benchmarks \cite{WahCUB_200_2011}. In addition, it would be informative to evaluate on other modalities (e.g.\ audio \cite{avmnist}) as well as multi-modal regression and reinforcement learning tasks. 






\clearpage
\begin{ack}

MJ and SM acknowledge funding from EPSRC Centre for Doctoral Training in Autonomous Intelligent Machines and Systems (Grant No: EP/S024050/1). MJ also acknowledges funding from AWS in collaboration with the Oxford-Singapore Human Machine Collaboration initiative. 

Work done partially while at University College London. We would also like to thank Yihong Chen and Tim Rockt\"aschel for initially facilitating the project.

\end{ack}

\printbibliography

\section*{Checklist}

\begin{enumerate}

\item For all authors...
\begin{enumerate}
  \item Do the main claims made in the abstract and introduction accurately reflect the paper's contributions and scope?
   \answerYes{}
  \item Did you describe the limitations of your work?
    \answerYes{See Section \ref{sec:conc}}
  \item Did you discuss any potential negative societal impacts of your work?
    \answerYes{See Broader Impacts section (Appendix \ref{sec:impact}.}
  \item Have you read the ethics review guidelines and ensured that your paper conforms to them?
    \answerYes{}
\end{enumerate}

\item If you are including theoretical results...
\begin{enumerate}
  \item Did you state the full set of assumptions of all theoretical results?
    \answerNA{}
	\item Did you include complete proofs of all theoretical results?
   \answerNA{}
\end{enumerate}

\item If you ran experiments...
\begin{enumerate}
  \item Did you include the code, data, and instructions needed to reproduce the main experimental results (either in the supplemental material or as a URL)?
    \answerYes{\url{https://github.com/s-a-malik/multi-few}}
  \item Did you specify all the training details (e.g., data splits, hyperparameters, how they were chosen)?
    \answerYes{See Section \ref{sec:setup} and Appendix \ref{sec:hypers}}
	\item Did you report error bars (e.g., with respect to the random seed after running experiments multiple times)?
    \answerYes{See Table \ref{tab:results} and Figure \ref{figure:fumi-vs-maml}}
	\item Did you include the total amount of compute and the type of resources used (e.g., type of GPUs, internal cluster, or cloud provider)?
    \answerYes{See Appendix \ref{sec:hypers}}
\end{enumerate}

\item If you are using existing assets (e.g., code, data, models) or curating/releasing new assets...
\begin{enumerate}
  \item If your work uses existing assets, did you cite the creators?
    \answerYes{See Section \ref{sec:data} and Appendix \ref{sec:appdata}}
  \item Did you mention the license of the assets?
    \answerYes{See Appendix \ref{sec:appdata}}
  \item Did you include any new assets either in the supplemental material or as a URL?
    \answerYes{\url{https://doi.org/10.5281/zenodo.6703088}}
  \item Did you discuss whether and how consent was obtained from people whose data you're using/curating?
   \answerYes{See Appendix \ref{sec:appdata}}
  \item Did you discuss whether the data you are using/curating contains personally identifiable information or offensive content?
    \answerYes{See Appendix \ref{sec:appdata}}
\end{enumerate}

\item If you used crowdsourcing or conducted research with human subjects...
\begin{enumerate}
  \item Did you include the full text of instructions given to participants and screenshots, if applicable?
   \answerNA{}
  \item Did you describe any potential participant risks, with links to Institutional Review Board (IRB) approvals, if applicable?
   \answerNA{}
  \item Did you include the estimated hourly wage paid to participants and the total amount spent on participant compensation?
    \answerNA{}
\end{enumerate}

\end{enumerate}

\appendix

\section{Broader Impact}
\label{sec:impact}

In this work we seek to develop better methods for few-shot learning. Few-shot learning has the potential to democratise access to powerful machine learning methods as it enables their usage in resource-constrained environments and minority groups which may not be well-represented in datasets which have largely been curated in western cultures. However, it could also have potential negative implications, for example, it could be used in facial recognition, reducing the privacy of individuals. 

The dataset we release with this work (iNat-Anim) was developed to help evaluate multi-modal few-shot learning. It consists of less noisy, short textual descriptions than previous works \cite{WahCUB_200_2011}. This enables method development in the field with smaller models and therefore it could reduce the environmental impact of training large models for research purposes. While we inspected the data as much as possible, we have not checked all of the descriptions obtained from the Animalia website. There always remains a risk with using web-scraped data that harmful or biased descriptions could be present, and as such the data must be used with care.

\section{Implementation Details}
\label{sec:hypers}

All models were implemented in \textit{PyTorch}
\cite{pytorch}. Each training run took 1.5 to 3 hours on a single free-tier \textit{Google Colaboratory} GPU. We used BERT \cite{bert} (\texttt{bert-base-uncased}) from the \textit{Hugging Face} library \cite{wolf2019huggingface} as the pre-trained text encoder for all models. We followed the standard pre-processing routine for BERT, which involved truncating descriptions that were longer than the maximum sequence length for the model. We use ResNet-152 \cite{he2016deep} as the pre-trained image encoder for all models. This has a feature dimension of 2048. The final layer (head) was fine-tuned but all other parameters were frozen during training.

The \textit{Torchmeta} library \cite{torchmeta} was used to construct meta-splits for the dataset. We used a 60:20:20 train:validation:test class splitting. Due to computational restrictions we could not perform extensive hyperparameter tuning. Instead, hyperparameters were chosen via simple heuristics that maximized accuracy on the validation split, using suggestions from the literature as starting points. The validation split was also used to select the best model checkpoint using the validation loss. 
Our code has been open-sourced\footnote{\url{https://github.com/s-a-malik/multi-few}}. Poignant hyperparameters were as follows:

\paragraph{AM3/Prototypical Networks}

We set the prototype dimension to 512, and used a single hidden layer neural network with hidden dimension 512 for each of the $g$ and $h$ networks. We found that the number of tasks per batch of 5, 3, 2 and 1 and query set size during training of 10, 8, 8 and 8 worked well heuristically for 1, 3, 5 and 10 shots respectively. In all cases, we used the Adam optimizer \cite{adam} with learning rate $0.001$. Additionally, we used dropout \cite{srivastava2014dropout} with $p=0.2$ and an L2 weight decay of 0.0005 to prevent overfitting. For the zero-shot AM3 model, we forced $\lambda=0$ to remove dependence of the prototype on the support images. For the uni-modal prototypical network baseline \cite{protonets}, we simply used our AM3 implementation but manually forced $\lambda=1$, which removes the dependence of the prototype on the text.  

\paragraph{FuMI/MAML} 
We use 4 tasks per batch, and a query set size during training of 32 for all shots. For both FuMI and MAML, the model consisted of three fully-connected layers, with hidden layer widths of 256 and 64. The FuMI hypernetwork also consisted of two fully-connected layers, with a hidden layer width of 256. A dropout rate of $p=0.25$ was used. Again, we used the Adam optimizer with learning rate 0.00003 and an L2 weight decay of 0.0005 for outer-loop training. For the inner-loop, a step size of 0.01 was used, with 5 training updates on the support set at meta-train time. At meta-test time, 50, 50, 100 and 100 inner-loop updates were performed on the support set, for 1, 3, 5 and 10 shots respectively.


\section{iNat-Anim Details}
\label{sec:appdata}

\begin{figure*}[ht]
    \centering
    \includegraphics[width=\linewidth]{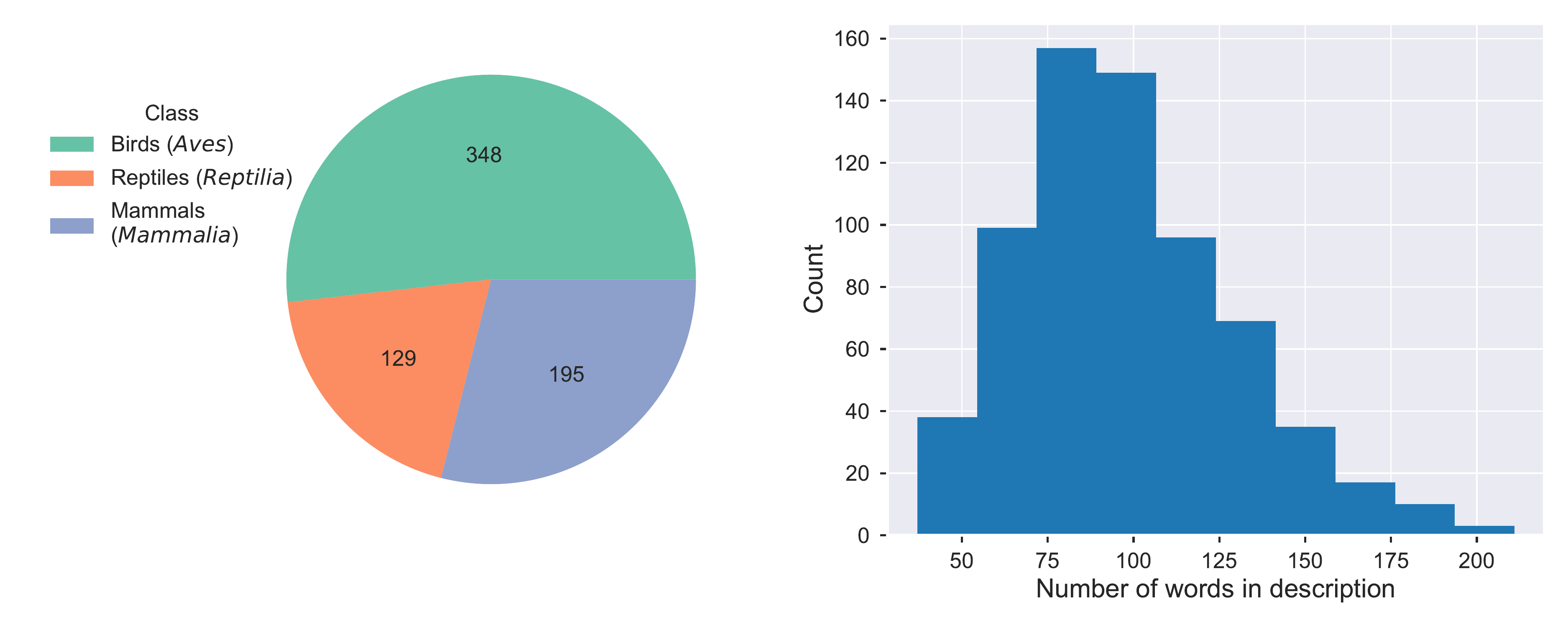}
    \caption{The left pane shows the distribution of species in iNat-Anim across birds, reptiles and mammals. The right pane shows a histogram of the number of words in each description of each species.}
    \label{figure:zanim-stats}
\end{figure*}

The images are a subset of the images from the iNaturalist 2021 CVPR challenge \cite{inat21} and have been augmented with textual descriptions of each species from Animalia \cite{animalia}, an online animal encyclopedia. Full permission for website scraping and dataset publication was obtained from the owners of Animalia prior to release of the dataset.
We place a \textit{CC BY-NC 4.0}\,\footnote{\url{https://creativecommons.org/licenses/by-nc/4.0/}} licence on the textual descriptions scraped from Animalia, whilst retaining the per-image licensing from the relevant subset of iNaturalist. The descriptions are curated by the website owners and we manually inspected a significant proportion for quality. To best of our knowledge, the descriptions do not contain any personally identifiable or offensive material. Figure \ref{figure:zanim-stats} shows the distributions of classes and description lengths across the dataset.

\end{document}